\documentclass[conference]{IEEEtran}
\usepackage{amsmath, booktabs, subcaption, graphicx, amssymb, algorithm, verbatim}
\usepackage[noend]{algpseudocode}

\ifCLASSOPTIONcompsoc

  \usepackage[nocompress]{cite}
\else
  \usepackage{cite}
\fi
\ifCLASSINFOpdf
\else

\fi

\begin{document}

\title{Measuring Inter-group Agreement on zSlice Based General Type-2 Fuzzy Sets}

\author{\IEEEauthorblockN{Javier Navarro and Christian Wagner}
\IEEEauthorblockA{Lab for Uncertainty in Data and Decision Making (LUCID),\\School of Computer Science, University of Nottingham, Nottingham, UK\\
Email: francisco.navarro@nottingham.ac.uk\\ christian.wagner@nottingham.ac.uk}}

\maketitle
\begin{abstract}
Recently, there has been much research into modelling of uncertainty in human perception through Fuzzy Sets (FSs). Most of this research has focused on allowing respondents to express their (\textit{intra}) uncertainty using intervals. Here, depending on the technique used and types of uncertainties being modelled different types of FSs can be obtained (e.g., Type-1, Interval Type-2, General Type-2). Arguably, one of the most flexible techniques is the Interval Agreement Approach (IAA) as it allows to model the perception of all respondents without making assumptions such as outlier removal or predefined membership function types (e.g. Gaussian). A key aspect in the analysis of interval-valued data and indeed, IAA based agreement models of said data, is to determine the position and strengths of agreement across all the sources/participants. While previously, the Agreement Ratio was proposed to measure the strength of agreement in fuzzy set based models of interval data, said measure has only been applicable to type-1 fuzzy sets. In this paper, we extend the Agreement Ratio to capture the degree of inter-group agreement modelled by a General Type-2 Fuzzy Set when using the IAA. This measure relies on using a similarity measure to quantitatively express the relation between the different levels of agreement in a given FS. Synthetic examples are provided in order to demonstrate both behaviour and calculation of the measure. Finally, an application to real-world data is provided in order to show the potential of this measure to assess the divergence of opinions for ambiguous concepts when heterogeneous groups of participants are involved.
\end{abstract}

\IEEEpeerreviewmaketitle

\section{Introduction}

There is a growing body of literature within the Fuzzy Set (FS) community and in particular within the \textit{Computing With Words} area \cite{Zadeh1996} that is based on the premise that ``words mean different things to different people'' \cite{Mendel2007a}. It is this premise that has motivated the development of robust techniques for modelling and reasoning \cite{Feilong2008, Coupland2010, Wu2012a,Wagner2014,Hao2016} using a number of different types of FS based models in respect to inter-individual uncertainty associated with terms.

Most of the existing approaches for modelling the variation in perception through FSs employ Interval-Valued (IV) data as the sourced standard for enabling the capture of information of imprecise nature. Some of the recent examples of these modelling methods are: the Interval Approach \cite{Feilong2008}, the Enhanced Interval Approach \cite{Coupland2010,Wu2012a}, the Hao-Mendel Approach \cite{Hao2016} and the Interval Agreement Approach (IAA) \cite{Wagner2014}.

While the aforementioned approaches and their generated FS based models of Linguistic Terms (LTs) have been found useful to enable computing with and understanding of key LTs, it is only the Interval Agreement Approach \cite{Wagner2014,Miller2012,Wagner2013b} that allows the modelling of IV data with minimal assumptions on the distributions of the generated FS models without requiring data pre-processing (and the potential omission of a large amount of the original data). Moreover, the IAA provides the flexibility to separately model different types of uncertainty, namely \textit{inter}- and \textit{intra}-source uncertainty. Further, its use on representation and analysis of crowd-sourced data has shown great promise in a number of studies \cite{Havens2010, Wallace2016, McCulloch2015, Miller2014, AlFarsi2017, Wagner2013b} in which the modelling of \textit{inter}- and/or \textit{intra}- source uncertainty is required to provide insight into a given concept of interest. In this sense, a key aspect of generating such quantitative models of the meaning of words/LTs, is the assessment of how clearly the resulting model reflects the meaning of a LT held by \emph{different groups} of people when the individual perception of each respondent is expressed using an interval.

In this paper, we continue exploring the use of IAA based modelling and IV data capture of human-centred data as initially shown in previous works \cite{Navarro2016a,Navarro2016b} to analyse agreement based models. In particular, we introduce an extension of a previously proposed index (the agreement ratio) for measuring, for the first time, the variation of perceptions among different groups of respondents (i.e., inter-group agreement) when using zSlice based General Type-2 Fuzzy Sets (zGT2 FSs) \cite{Wagner2010a}.

The paper is structured as follows: Section \ref{bck} provides background on the \textit{zSlice} based representation of GT2 FSs and the IAA as well as the previously proposed index i.e., the original Agreement Ratio for intra-group agreement (and type-1 FSs). In Section \ref{agrRatGT2}, we introduce the novel index for assessing the extent of inter-group agreement modelled by zGT2 FSs. Section \ref{res}, presents demonstrations of the proposed index using both synthetic and real world data from a case study. Finally, Section \ref{conc} summarises the contributions of this paper.


\section{Background} \label{bck}
This Section provides detail on key background material for the paper. Section \ref{sec:gt2fs} presents a brief overview of the zSlice based representation of GT2 FSs. This is followed by a description of the Interval Agreement Approach in Section \ref{sec:iaa} given that it serves as the basis for modelling \textit{inter}- and \textit{intra}-source uncertainty through FSs. Finally, we review the Agreement Ratio for IAA T1 FS models in Section \ref{agrRatT1}.

\subsection{General Type-2 Fuzzy Sets based on zSlices} \label{sec:gt2fs}
General Type-2 Fuzzy Sets (GT2 FSs)\cite{Zadeh1975A} are FS models based on representing the membership grade of its elements to a (T2) FS using a secondary grade which itself is a T1 FS (i.e., $\mu_{X}(x)\in [0,1]$). Due to their greater computational cost several representations simplifying their concepts and operations have been proposed \cite{Wagner2008, Wagner2010a, Mendel2009, Coupland2007}.

In this paper, we focus our descriptions on the zSlice based representation \cite{Wagner2008} (or alpha-planes \cite{Liu2008}) of GT2 FSs as the index proposed in Section \ref{agrRatGT2} is based on a technique (the IAA) built upon this representation.


General Type-2 Fuzzy Sets based on \textit{z}Slices (zGT2 FSs) \cite{Wagner2008} can be seen as a proportional `segmentation' of a GT2 FS in the secondary grade (here referred as \(z\) dimension). This segmentation results in a `slice' or zSlice, each of which, is assigned a height $z_j$. Each \textit{z}Slice, is equivalent to an Interval Type-2 FS with the exception that its secondary grade (3rd dimension) is equal to $z_j$ instead of being fixed to 1, where $0\leq z_j \leq 1$. Thus, each \textit{z}Slice $\tilde{Z}_j$ can be described as follows 
\begin{equation}
\tilde { Z }_j =\sum _{ x\in \mathcal{X} }{ \sum _{ u\in \left[ { \underline { u }  }_{ x },{ \overline { u }  }_{ x } \right]  }{ z_j/(x,u) }  } 
\end{equation}
Using this representation, a GT2 FS can then be approximated and represented as a collection of $N$ \textit{z}Slices such that:
\begin{equation}
\tilde{X}={\sum}_{j=1}^{N}\tilde{Z}_j
\end{equation}
where $N$ is the number of \textit{z}Slices. 

We now move to define a key method used in this paper for modelling of IV data using Type-1 and Type-2 FSs, the Interval Agreement Approach.

%
%

\subsection{Interval Agreement Approach}\label{sec:iaa}
The IAA \cite{Wagner2014} is designed to model and represent the agreement across a group (or groups) of people who provided their responses w.r.t. a specific matter in the form of intervals. Depending on the type of intervals used (i.e., crisp or uncertain intervals) and number of sources to distinguish from (e.g., groups of people), several types of FSs (Type-1, Interval Type-2 and zSlices based General Type-2 FSs \cite{Wagner2010a}) can be generated using the IAA. In this paper we consider the case of using crisp intervals (i.e., no uncertainty about the interval endpoints), therefore the definitions on the IAA are focused on T1 and GT2 FSs. What follows in this section is an account of the selected IAA sub methods for generating FS models from IV data.

\subsubsection{T1 FS based IAA modelling of intra/inter-source uncertainty}
T1 FSs are obtained using the IAA when crisp intervals and either \textit{inter}- or \textit{intra}-source uncertainty are modelled in the primary degree of membership by combining multiple intervals. The following describes this process:

Consider $ n $ (closed) intervals $\overline { x }_i = [{ x }^{ - }_{i},{ x }^{ + }_{i}] $, $ i\in \left\{ 1,...,n \right\} $ to be modelled as a T1 FS $ X $. The membership function of $ X $ (denoted by ${\mu}_{X}$) is described as follows:
\begin{equation}\label{eq:iaaSets}
\begin{aligned}
\mu (X)={} & { { y }_{ 1 } }/{ \bigcup _{ { i }_{ 1 }=1 }^{ n }{ { \bar { x }  }_{ { i }_{ 1 } } }  }\\
& +{ { y }_{ 2 } }/\left( { \bigcup _{ { i }_{ 1 }=1 }^{ n-1 }{ \bigcup _{ { i }_{ 2 }={ i }_{ 1 }+1 }^{ n }{ \left( { \bar { x }  }_{ { i }_{ 1 } }\cap { \bar { x }  }_{ { i }_{ 2 } } \right)  }  }  } \right) \\
&+{ { y }_{ 3 } }/\left( { \bigcup _{ { i }_{ 1 }=1 }^{ n-2 }{ \bigcup _{ { i }_{ 2 }={ i }_{ 1 }+1 }^{ n-1 }{ \bigcup _{ { i }_{ 3 }={ i }_{ 2 }+1 }^{ n }{ \left( { \bar { x }  }_{ { i }_{ 1 } }\cap { \bar { x }  }_{ { i }_{ 2 } }\cap { \bar { x }  }_{ { i }_{ 3 } } \right)  }  }  }  } \right) \\
&+\cdots \\
& +{ { y }_{ n } }/\left( { \bigcup _{ { i }_{ 1 }=1 }^{ 1 }{ \cdots \bigcup _{ { i }_{ n }=n }^{ n }{ \left( { \bar { x }  }_{ { i }_{ 1 } }\cap \dots \cap { \bar { x }  }_{ { i }_{ n } } \right)  }  }  } \right) ,
\end{aligned}
\end{equation}
where ${y}_{i}=\frac{i}{n}$ and $/$ refers to the common notation of membership, not division. Thus, the main idea of the IAA is for the FS to model the overall information across the universe of discourse while reflecting the extent of agreement among $n$-tuples of intervals (IV responses) throughout the $y$ axis.

\subsubsection{GT2 FS based IAA modelling of intra- and inter-source uncertainty}
General Type-2 FS based on zSlices are obtained when two types of source uncertainty (e.g., inter-source and inter-group) are being modelled through the primary and secondary degrees of membership. 

Similarly, consider $ N $ T1 FSs ${ X }_i$, $ i\in \left\{ 1,...,N \right\} $ to be modelled as a GT2 FS $\tilde{X}$. The $N$ zSlices of $\tilde{X}$ and their secondary membership are described in a similar manner as in the T1 case but, in this case the secondary membership values are defined by ${z}_{I}=\frac{I}{N}$ whereas the zSlices are defined by the union of intersections of $1,\dots,N$ tuples of source FSs (not intervals as in the previous case).

\subsection{Agreement Ratio for IAA generated T1 FS models} \label{agrRatT1}
Let $X$ be a FS modelled from a set of intervals $ \overline { x } = \left\{ \overline{ x }_1,...,\overline{ x }_n \right\} $ where $ \overline{x}_i = \left[ { x }^{ - }_{i},{ x }^{ + }_{i} \right]$ such that $\overline{x}_i=\left\{ x\in \mathbb{R}|{ x }^{ - }_i \le x\le { x }^{ + }_i  \right\} $. An agreement ratio $\gamma$ is obtained for a T1 FS with the following equation:

\begin{equation}\label{eq:agrRat}
\gamma \left( X \right) =\frac {  \left( { \sum _{ \alpha =2 }^{ n }{ { y }_{ \alpha  }\left( \frac { { \left| X_{ \alpha  } \right|  } }{ { \left| { X_{ \alpha -1 } } \right|  } }  \right)  }  } \right)  }{ \sum _{ \alpha =2 }^{ n }{ { y }_{ \alpha  } } } 
\end{equation}
where $ 0\le \gamma\le 1$ and ${y}_{\alpha}=\frac { \alpha }{ n }$ weights the relation between immediate agreement $y$ levels in question. It can be noticed that, as per design of the IAA, the $\alpha$-cut ${ { X }_{ \alpha=1 } }$ is directly related to the union of all source intervals, ${ { X_{ \alpha =2 } } }$ is equal to the cardinality of the union of all tuples of intervals where there is at least 2 joint intervals, and so on. Given that it is true that ${{ X_{ \alpha } } }\subseteq {{ X_{ \alpha -1 } } }, \forall \alpha \geq 2$ then the more \emph{similar} their cardinality is the more \textit{inter}-source agreement is present, hence the division operator\footnote{The rationale of doing so is due to the use of the Jaccard similarity measure \cite{Jaccard1908} which is simplified to the case of $\frac{|A|}{|B|}$ when $A \subseteq B$.}. 

\begin{figure}[ht]
	\includegraphics[scale=0.2]{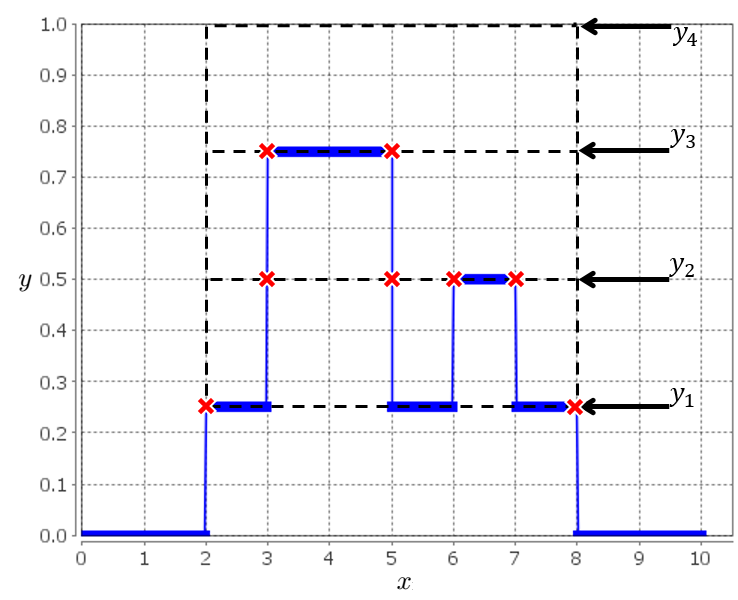}
	\centering
	\caption{Fuzzy set generated from 4 intervals: \({ \overline { D }  }_{ 1 }=\left[ 2,5 \right] \) and \({ \overline { D }  }_{ 2 }=\left[ 3,5 \right] \), \({ \bar { D }  }_{ 3 }=\left[ 6,8 \right] \) and \({ \overline { D }  }_{ 4 }=\left[ 3,7 \right] \)}
	\label{fig:fsB}
\end{figure}

Note that this approach can be applied to both convex and non-convex FSs as ${ \left| { X_{ \alpha } }  \right|  }$ only considers the sum of sizes of \emph{existing} intersections of $ \alpha$ intervals and thus, the `gaps' between those intervals are not considered in calculations such as shown in Fig. \ref{fig:fsB}.

Finally, the overall summation is divided by the sum of `weights' $y_\alpha$ so the final ratio is normalised to a number in the range [0,1]. As can be seen in \eqref{eq:agrRat}, these $y_\alpha$ terms act as indicators of the level of agreement weighting overlapping $\alpha$ -cuts in proportion to the number of intervals considered in the intersection operation. 

In order to compute the cardinality of the ${ \left| { X_{ \alpha } } \right| }$ intersections, Algorithm \ref{alg:alpha} can be used which is based on `discretisations' of resulting alpha cuts. 

\begin{algorithm}
\caption{Calculation of $\alpha$-cuts lengths}\label{alg:alpha}
\begin{algorithmic}[1]
\Procedure{AlphaLength}{$\alpha$, $X$}\Comment{Returns: The length of the existing intervals at a given $\alpha$-cut.}
\State $l\gets 0$, $r\gets 0$ The points where an $\alpha$-cut `intersects' the FS.
\State $ n\gets $ \# of discretisations
\State discretise($x$)\Comment{discretise domain $x_1,...,x_i,...x_n$}
\State $b\gets $false\Comment{Boolean for detection of intervals}
\For{$i=1$ to $n$}
\State $y_i\gets {\mu}_{X}(x_i)$ \Comment{Get the membership value at $x_i$}
\If{$y_i<\alpha$}
\State $y_i\gets 0$
\If{b = true}\Comment{If we detected a `fall' below the alpha level}
\State $r\gets {x}_{i-1}$ 
\State addCut$(l,r)$\Comment{Add the detected interval}
\EndIf
\State $b\gets$ false
\Else
\If {$b=$false}
\State $l\gets{x}_{i}$
\EndIf
\State $b\gets true$
\EndIf
\EndFor
\If {$b= true$}\Comment{If there is an interval ongoing at the end of discretisations.}
\State $r\gets x_{n}$
\State addCut$(l,r)$
\EndIf
\For{each $\alpha$Cut $j$}
\State $length\gets length + (r_{j}-l_{j})$\Comment{Interval Size}
\EndFor
\State \textbf{return} $length$
\EndProcedure
\end{algorithmic}
\end{algorithm}

This section has described the IAA technique for the cases of modelling T1 and GT2 FSs from crisp intervals including the agreement ratio for the case of T1 FSs. What follows describes the proposed index for measuring the inter-group agreement when modelling perception through a zGT2 FS.

\section{Agreement Ratio for IAA GT2 FS models} \label{agrRatGT2}
This Section presents an index called the \textit{inter}-group Agreement Ratio for calculating a useful value for the analysis of GT2 FSs models generated using the IAA.

Just as described in Section \ref{sec:iaa}, the Interval Agreement Approach has been proposed as a means of modelling the agreement of IV data from different sources using Fuzzy Sets. Depending on the application and intervals used, the IAA can model different types of uncertainty associated to linguistic terms through the primary and secondary membership functions of zSlices based General Type-2 Fuzzy Sets \cite{Wagner2008}.

For example, one can argue that an expression such as `A little bit difficult' can not only be uncertain to some extent for one self (over time and across different contexts for example) but also perceived in varying ways by different people or groups of so with distinct knowledge/background.
\begin{figure}[!ht]
	\centering
	\begin{subfigure}[b]{0.21\textwidth}
		\includegraphics[width=\textwidth]{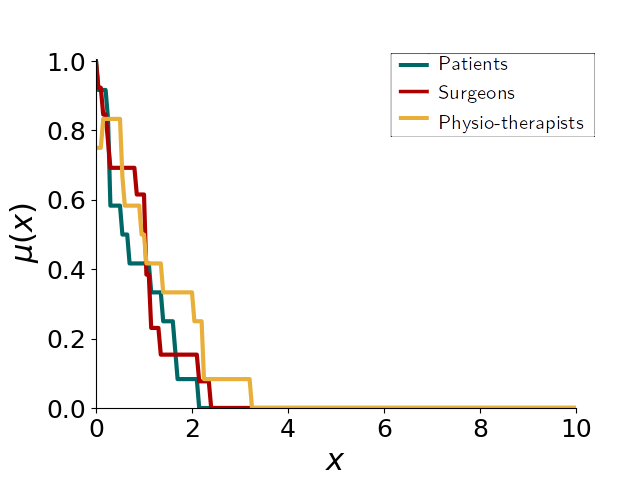}\caption{Impossible to do}
		\label{subfig:splitITD}
	\end{subfigure}
	\begin{subfigure}[b]{0.21\textwidth}
		\includegraphics[width=\textwidth]{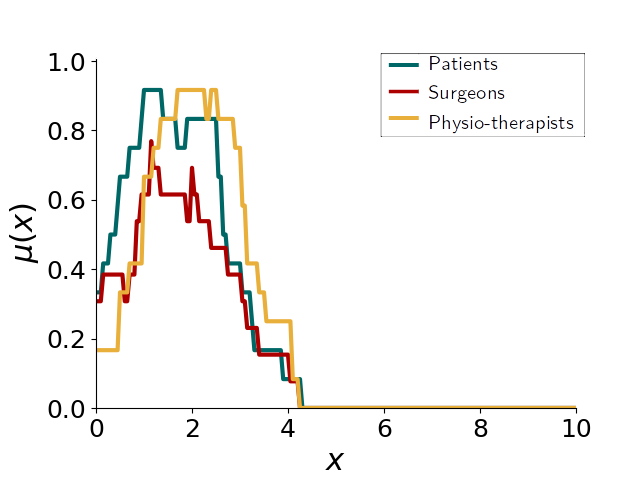}\caption{Extremely difficult}
		\label{subfig:splitED}
	\end{subfigure}
	~
	\begin{subfigure}[b]{0.21\textwidth}
		\includegraphics[width=\textwidth]{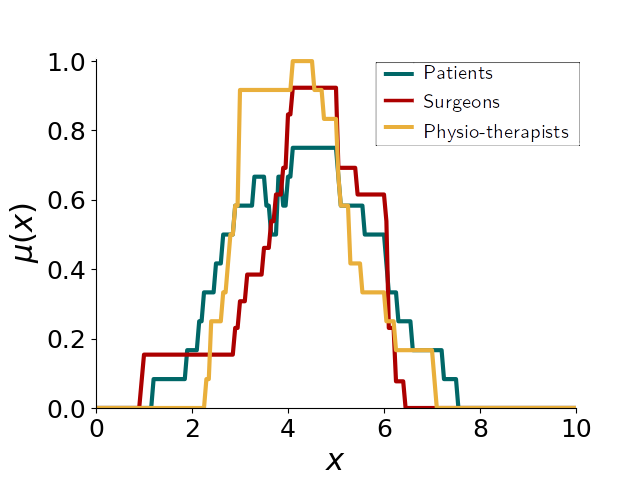}\caption{Moderately Difficult}
		\label{subfig:splitMD}
	\end{subfigure}
	\begin{subfigure}[b]{0.21\textwidth}
		\includegraphics[width=\textwidth]{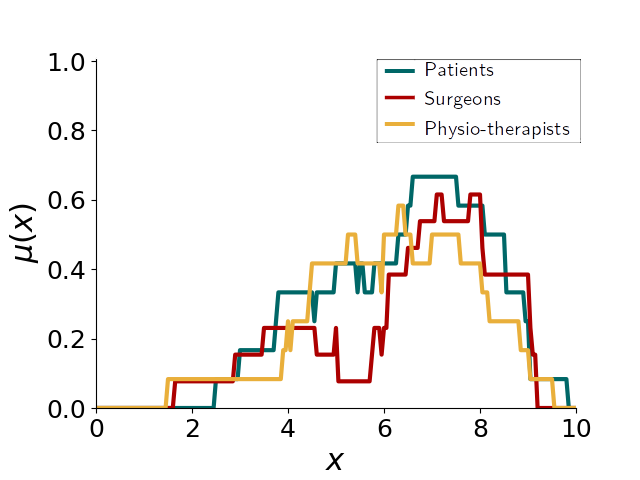}\caption{A little bit difficult}
		\label{subfig:splitALBD}
	\end{subfigure}
	~
	\begin{subfigure}[b]{0.21\textwidth}
		\includegraphics[width=\textwidth]{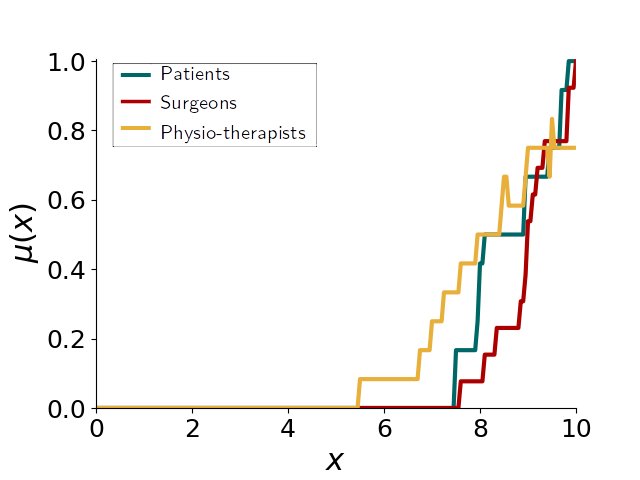}\caption{Not at all difficult}
		\label{subfig:splitNAAD}
	\end{subfigure}
	\caption{T1 FSs generated from Patients, Physiotherapists and Surgeons contrasting their perception on the linguistic descriptors: \textit{ITD}, \textit{ED}, \textit{MD}, \textit{ALBD}, and \textit{NAAD}. The inter-group variation of these source T1 FSs are in turn modelled by GT2 FSs and shown in Figure \ref{fig:gt2LTs}}
	\label{fig:T1LTs}
\end{figure}
Figure \ref{fig:T1LTs} illustrates this point with 5 Linguistic Terms modelled using the IAA representing the corresponding perception for three groups of people (i.e., patients, physiotherapists and surgeons) who provided an IV response representing their perception on a scale. Note that as there are 3 groups of people involved with varying perceptions, a GT2 with 3 zSlices can be used to visualise and separate inter-individual and inter-group agreement through the $y$ and $z$ dimensions of the generated FS respectively. In this sense, the inter-group agreement can be inferred from the FS by looking at how much the zSlices `shrinks' from the first to the last zSlice. While this visual representations are informative it can be difficult to assess objectively how well perceived is a LT among different groups of people.

As proposed by Wagner et al. \cite{Wagner2014}, and explained in Section \ref{sec:iaa}, IAA generated GT2 FSs model the overall perception by reflecting the agreement in the source T1 FSs through the secondary membership. This secondary membership effectively weights `areas' accounting for the extent of overlapping between $N$ T1 FSs such that the higher the similarity of subsequent zSlices is, the more (inter-group) agreement exists. To illustrate this reasoning, consider two contrasting cases in which two T1 FSs ($N=2$) obtained from two groups of people are used to create a GT2 FS using the IAA:
\begin{itemize}
    \item Two identical FSs. An agreement ratio should be equal to 1 given that both groups have totally equal perception represented by two equal T1 FSs (see Fig. \ref{subfig:totalAg}). 
    \item Two disjoint FSs. An agreement ratio should be equal to 0 given that there are no overlapping regions (see Fig. \ref{subfig:nullAg}). In this case, the zSlice associated to the intersection operation of both source FSs is not present.
\end{itemize}
From these initial cases, it can be seen that at the highest level of secondary membership (i.e., $z_N$), there might be a \textit{z}Slice representing the regions where the $N$ source (group related) FSs agree to a $y$ extent, at the ${z}_{N-1}$ level, there might be certain overlapping where at least $N-1$ groups agree and so on. Thus, following a similar reasoning as in Section \ref{agrRatT1} if the \textit{similarity} between immediate zSlices is quantitatively expressed, then a ratio representing their overall agreement can be obtained. This relation can be obtained using a similarity measure as a basis\footnote{Again, in this paper we consider the Jaccard similarity measure \cite{Jaccard1908} as a basis to quantify this relation but others can also be explored.}. Moreover, these relations between immediate \textit{z}Slices can be weighted in order to assign greater relevance to higher secondary membership values, that is, source FSs where more inter-group agreement is present. From these base cases and reasoning we can proceed to generalise the following measure for calculating an inter-group agreement ratio.

Let $\tilde{X}$ be a GT2 FS modelled from a set of IAA generated T1 FSs $\{X_1,\dots,X_N\}$ representing the agreement across different groups of people. An agreement ratio $\gamma$ ($0 \le \gamma \le 1$) is obtained from $\tilde{X}$ using the following expression:
\begin{equation}\label{eq:agrRatGT2}
\gamma \left( \tilde { X }  \right) =\frac { \left( { \sum _{ j=2 }^{ N }{ { z }_{ j }\left( \mathbf{S}({ \tilde { Z }  }_{ j },{ \tilde { Z }  }_{ j-1 })\right)  }  } \right)  }{ \sum _{ j=2 }^{ N }{ { z }_{ j } }  } 
\end{equation}

where $z_{ j }=\frac { j }{ N } $ and $ \mathbf{ S }({ \tilde { Z }  }_{ j },{ \tilde { Z }  }_{ j-1 }) $ is the simplified Jaccard similarity measure applied to the \textit{z}Slices ${ \tilde { Z }  }_{ j }$ and ${ \tilde { Z }  }_{ j-1 }$. It is important to consider that as we are considering crisp intervals and therefore T1 FSs are created for each group of people then the resultant \textit{z}Slices are IT2 FSs in which both Upper and Lower MFs are equal. Also, note that as we compare immediate \textit{z}Slices it is always true that $\tilde { Z }_{j}$ is contained in $\tilde { Z }_{j-1}$, i.e., $\tilde {Z}_{j}\subseteq \tilde { Z }_{j-1}$ thus, the similarity $ \mathbf{S}({ \tilde { Z }  }_{ j },{ \tilde { Z }  }_{ j-1 }) $ can be simplified as follows:
\begin{equation}\label{eq:jacSim1}
\begin{aligned}
\mathbf{S}({ \tilde { Z }  }_{ j },{ \tilde { Z }  }_{ j-1 })={} & \frac { { \tilde { Z }  }_{ j }\cap { \tilde { Z }  }_{ j-1 } }{ { \tilde { Z }  }_{ j }\cup { \tilde { Z }  }_{ j-1 } } =\frac { { \tilde { Z }  }_{ j } }{ { \tilde { Z }  }_{ j-1 } } \\
& =\frac { \sum _{ i=1 }^{ n }{ { \overline { \mu  }  }_{ \tilde { Z } _{ j } }(x_{ i }) } +\sum _{ i=1 }^{ n }{ { \underline { \mu  }  }_{ \tilde { Z } _{ j } }(x_{ i }) }  }{ \sum _{ i=1 }^{ n }{ { \overline { \mu  }  }_{ \tilde { Z } _{ j-1 } }(x_{ i }) } +\sum _{ i=1 }^{ n }{ { \underline { \mu  }  }_{ \tilde { Z } _{ j-1 } }(x_{ i }) }  } 
\end{aligned}
\end{equation}
This simplification on \eqref{eq:jacSim1} reduces the computations by using the membership values of T1 FSs to which we refer as ${ \mu }_{\tilde{ Z }}(x_i)$. This simplification is of great advantage as the similarity between each zSlice can be simply computed as shown in \eqref{eq:jacSim2}.
\begin{equation}\label{eq:jacSim2}
\mathbf{S}({ \tilde { Z }  }_{ j },{ \tilde { Z }  }_{ j-1 })=\frac { \sum _{ i=1 }^{ n }{ { { \mu  }  }_{ \tilde { Z } _{ j } }(x_{ i }) } }{ \sum _{ i=1 }^{ n }{ {  { \mu  }  }_{ \tilde { Z } _{ j-1 } }(x_{ i }) }  } 
\end{equation}
This section has introduced and described an index for measuring inter-group agreement when IV data (with crisp endpoints) is exploited to generate a T1 FS modelling each group (inter-participant) perception. The next part of this paper will provide evidence on the results of such measure when applied to real world data as well as numerical demonstrations with synthetic data.

\section{Experiments \& Discussion}\label{res}
Having described the proposed measure for inter-group agreement in GT2 FSs based on zSlices, the following subsections will demonstrate how this measure behaves when considering cases of \textit{total}, \textit{null} and \textit{moderate} inter-group agreement using simple synthetic cases. Lastly we present the application results of the measure using a real world case study with 3 groups of people. It is worth noting that as we are modelling IV data with `crisp' endpoints this leads to \textit{z}Slice based GT2 FSs where each \textit{z}Slice has equal upper and lower membership functions and hence, can be represented using T1 FSs as will be seen in the following examples.

\subsection{Numeric Examples with total, moderate and null inter-group agreement}\label{ss:ex1}
\begin{figure}[ht]
	\centering
    \begin{subfigure}[b]{0.2\textwidth}
		\includegraphics[width=\textwidth]{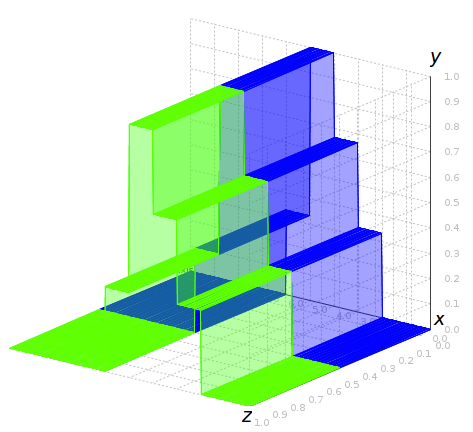}\caption{The GT2 FS $\tilde{T}$ generated from the FSs $T_1$ and $T_2$ showing total agreement.}
		\label{subfig:totalAg}
	\end{subfigure}
	\begin{subfigure}[b]{0.2\textwidth}
		\includegraphics[width=\textwidth]{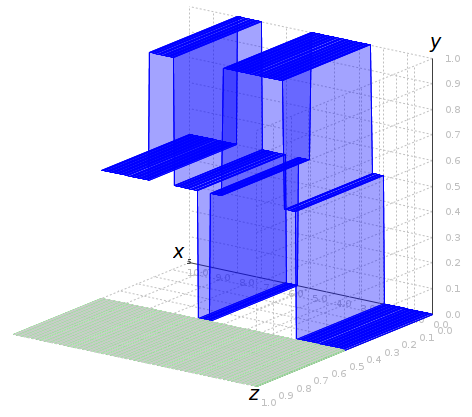}\caption{The GT2 FS $\tilde{N}$ generated from the FSs $N_1$ and $T_2$ showing null agreement.}
		\label{subfig:nullAg}
	\end{subfigure}
	~
    \begin{subfigure}[b]{0.2\textwidth}
		\includegraphics[width=\textwidth]{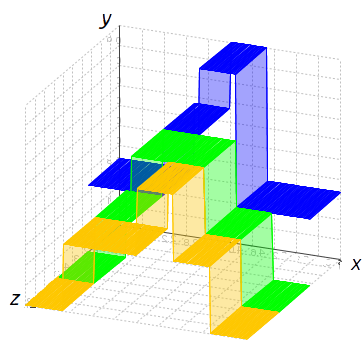}\caption{The GT2 FS $\tilde{M}$ generated from the FSs $M_1$, $M_2$ and $M_3$ showing moderate agreement.}
		\label{subfig:moderateAg}
	\end{subfigure}
	\caption{The GT2 FSs $\tilde{T}$, $\tilde{N}$ and $\tilde{M}$ illustrating three synthetic cases of \textit{Total}, \textit{Moderate} and \textit{Null} inter-group agreement from the intervals shown in Table \ref{tab:iaaGT2}.}
	\label{fig:iaaGT2}
\end{figure}

\begin{figure}[!ht]
	\centering
	\begin{subfigure}[b]{0.2\textwidth}
		\includegraphics[width=\textwidth]{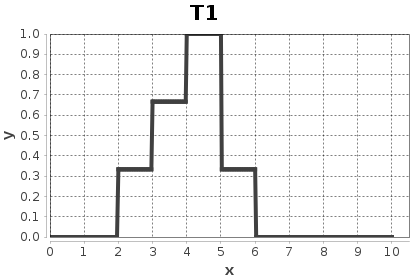}\caption{\textit{Total} agreement if $T_2$ is equivalent to $T_1$.}
		\label{subfig:total}
	\end{subfigure}
	\begin{subfigure}[b]{0.2\textwidth}
		\includegraphics[width=\textwidth]{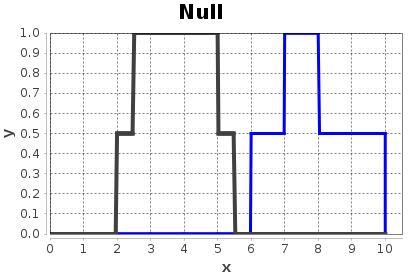}\caption{\textit{Null} inter group perception shown by $N_1$ and $N_2$.}
		\label{subfig:null}
	\end{subfigure}
	~
	\begin{subfigure}[b]{0.2\textwidth}
		\includegraphics[width=\textwidth]{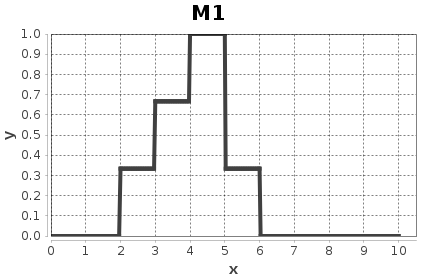}\caption{The T1 FS $M_1$ modelling the perception of group 1.}
		\label{subfig:M1}
	\end{subfigure}
    \begin{subfigure}[b]{0.2\textwidth}
		\includegraphics[width=\textwidth]{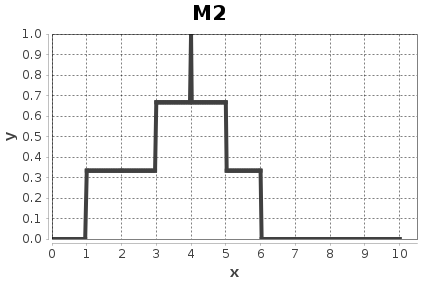}\caption{The T1 FS $M_2$ modelling the perception of group 2.}
		\label{subfig:M2}
	\end{subfigure}
    ~
	\begin{subfigure}[b]{0.2\textwidth}
		\includegraphics[width=\textwidth]{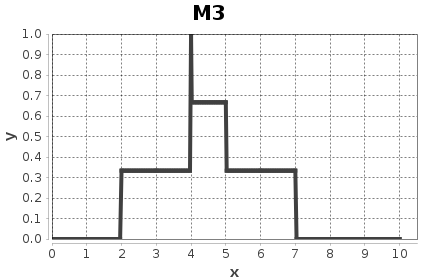}\caption{The T1 FS $M_3$ modelling the perception of group 3.}
		\label{subfig:M3}
	\end{subfigure}
    \begin{subfigure}[b]{0.2\textwidth}
		\includegraphics[width=\textwidth]{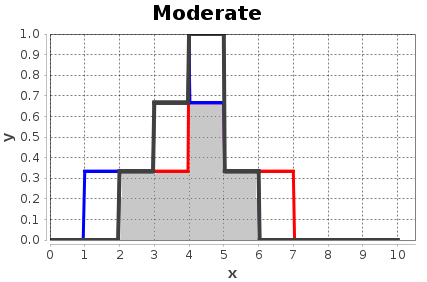}\caption{\textit{Moderate} inter group perception obtained from $M_1$, $M_2$ and $M_3$.}
		\label{subfig:Moderate}
	\end{subfigure}
	\caption{T1 FSs modelling the perception of different groups illustrating the cases of \textit{Total} (where $T_1$ is equivalent to $T_2$), \textit{Null} and \textit{Moderate} inter group agreement. Note that \textit{Moderate} inter group agreement is shown as the shaded area where all groups' FS models intersect.}
	\label{fig:synthAgGT2}
\end{figure}

\emph{Total agreement}. Consider the case of two groups of people, individuals of which provided the intervals shown in Table \ref{tab:iaaGT2} being modelled by the T1 FSs $T_1$ and \(T_2\). Note that if both FSs are identical thus, the inter-group agreement of the modelled GT2 FS \(\gamma (\tilde{T})\) (see Fig. \ref{subfig:totalAg}) is equal to 1 as shown in the following:
\begin{equation}\label{eq:agTotal}
\begin{aligned}
\gamma \left( \tilde { T }  \right) ={} &{ \left( \frac { 2 }{ 2 } \left( \mathbf{ S }({ \tilde { Z }  }_{ 2 },{ \tilde { Z }  }_{ 1 }) \right) \right)  }/{ (\cfrac { 2 }{ 2 } ) }\\
& ={ ( 1 ( 1 )  )}/{ 1 }=1.
\end{aligned}
\end{equation}
\emph{Null agreement}. Again, consider the case of two groups of people which provided the intervals presented in Table \ref{tab:iaaGT2} (described by the T1 FSs $N_1$ and $N_2$). These FSs clearly show that between these groups there is no agreement as their intersection is null. Thus, the inter-group agreement $\tilde{N}$ of the modelled GT2 FS shown in Fig. \ref{subfig:nullAg} is 0 as shown in the following expression: 
\begin{equation}\label{eq:agNull}
\begin{aligned}
\gamma \left( \tilde { N }  \right) ={} &{ \left( \frac { 2 }{ 2 } \left( \mathbf{ S }({ \tilde { Z }  }_{ 2 },{ \tilde { Z }  }_{ 1 }) \right) \right)  }/{ (\cfrac { 2 }{ 2 } ) }\\
& ={ ( 1 ( 0 )  )}/{ 1 }=0.
\end{aligned}
\end{equation}
\emph{Moderate agreement}. Let us now consider the GT2 FS $\tilde{M}$ depicted in Fig. \ref{subfig:moderateAg} generated by the intervals and T1 FSs ($M_1$, $M_2$ and $M_3$) described in Table \ref{tab:iaaGT2}. As can be seen, there is inter-group agreement to `some extent' as indicated by the existing $z_N$ zSlice representing the intersection of the three FSs. This inter-group agreement is measured by $\gamma(\tilde{M})$ so that it can be expressed as follows: 
\begin{equation}\label{eq:ex1}
\gamma \left( \tilde { M }  \right) =\frac { \left( \frac { 3 }{ 3 } \left( \mathbf{ S }({ \tilde { Z }  }_{ 3 },{ \tilde { Z }  }_{ 2 }) \right) +\frac { 2 }{ 3 } \left( \mathbf{ S }({ \tilde { Z }  }_{ 2 },{ \tilde { Z }  }_{ 1 }) \right)  \right)  }{ (\cfrac { 3 }{ 3 } +\cfrac { 2 }{ 3 } ) }
\end{equation}
Again, note that as intervals with crisp endpoints are used then each zSlice $\tilde{Z}_j$ has identical upper and lower membership functions as a result of using the IAA as modelling technique, therefore T1 FSs can be used to represent the different \textit{z}Slices as shown in Fig. \ref{fig:synthAgGT2} and expressed as follows:
\begin{equation*}
\tilde { Z } _{ 1 }={ \frac { 1 }{ 3 }  }/{ { \left( M_{ 1 }\cup M_{ 2 }\cup M_{ 3 } \right)  } }
\end{equation*}
\begin{equation*}
\tilde { Z } _{ 2 }={ \frac { 2 }{ 3 }  }/{ (\left( M_{ 1 }\cap M_{ 2 } \right) \cup \left( M_{ 1 }\cap M_{ 3 } \right) \cup \left( M_{ 2 }\cap M_{ 3 } \right) ) }
\end{equation*}
\begin{equation*}
\tilde { Z } _{ 3 }={ \frac { 3 }{ 3 }  }/{ { \left( M_{ 1 }\cap M_{ 2 }\cap M_{ 3 } \right)  } }
\end{equation*}

Having defined each zSlice, $ \gamma( \tilde { M } ) $ can be calculated by replacing the equation above in \eqref{eq:ex1} and using \eqref{eq:jacSim2} such that we obtain:
\begin{equation}
\gamma \left( \tilde { M }  \right) =\cfrac{ (\frac { \sum _{ i=1 }^{ n }{ { { \mu  } }_{ \tilde { Z } _{ 3 } }(x_{ i }) }  }{ \sum _{ i=1 }^{ n }{ { { \mu  } }_{ \tilde { Z } _{ 2 } }(x_{ i }) }  } +\frac { 2 }{ 3 } \left( \frac { \sum _{ i=1 }^{ n }{ { { \mu  } }_{ \tilde { Z } _{ 2 } }(x_{ i }) }  }{ \sum _{ i=1 }^{ n }{ { { \mu  } }_{ \tilde { Z } _{ 1 } }(x_{ i }) }  }  \right) ) }{ { 5 }/{ 3 } }\approx 0.772
\end{equation}
\begin{table}[htbp]
  \small
  \centering
  \caption{Intervals used to generate the GT2 FSs $\tilde{T}$, $\tilde{N}$ and $\tilde{M}$ representing cases of \textit{total}, \textit{null} and \textit{moderate} inter-group agreement respectively}
    \begin{tabular}{c|ccc|c}
    \toprule
    FS    &  \multicolumn{3}{c}{Intervals used} & $\gamma$\\
    \midrule
    $T_1$     & [2,5] & [4,5] & [3,6] & 0.5\\
    $T_2$     & [2,5] & [4,5] & [3,6] & 0.5\\
    \midrule
    $N_1$     & [2,5] & [2.5,5.5] &  & 0.714\\
    $N_2$     & [6,8] & [7,10] & & 0.25\\
    \midrule
    $M_1$     & [2,5] & [4,5] & [3,6] & 0.5\\
    $M_2$     & [1,4] & [3,5] & [4,6] & 0.16\\
    $M_3$     & [2,4] & [4,5] & [4,7] & 0.16\\    
    \bottomrule
    \end{tabular}%
  \label{tab:iaaGT2}%
\end{table}%

The results presented in these three synthetic cases are intended to illustrate the intuitive behaviour of the proposed measure as a tool for assessing how much agreement there is between groups of people with possibly different (uncertain) perceptions. As can be seen, not only in the simplest cases an intuitive result is obtained but also in the less obvious case, i.e., \textit{moderate}. We now turn to analyse the results obtained by the application of the presented measure to the data of our case of study, the IV TESS.

\subsection{Inter-group agreement in the IV TESS case}\label{ss:ex2}
\begin{figure}[!ht]
	\centering
	\begin{subfigure}[b]{0.2\textwidth}
		\includegraphics[width=\textwidth]{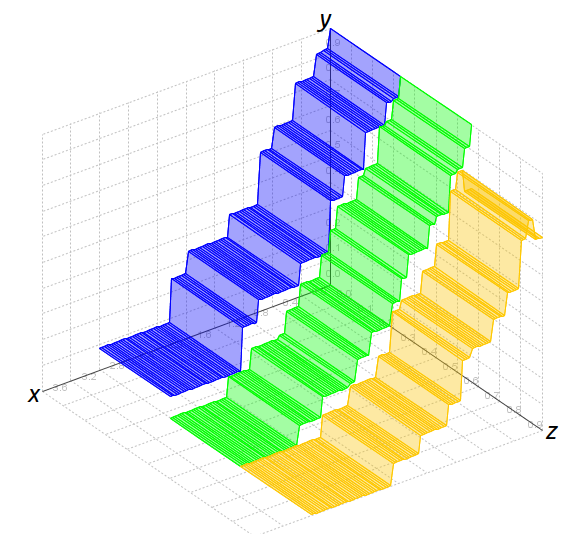}\caption{Impossible to do}
		\label{subfig:ITD}
	\end{subfigure}
	\begin{subfigure}[b]{0.2\textwidth}
		\includegraphics[width=\textwidth]{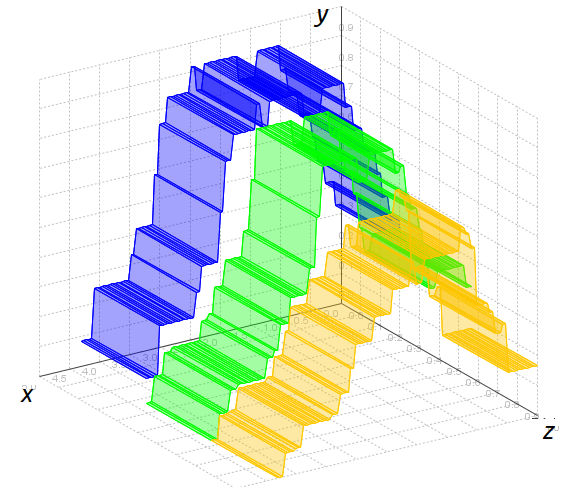}\caption{Extremely difficult}
		\label{subfig:ED}
	\end{subfigure}
	~
	\begin{subfigure}[b]{0.2\textwidth}
		\includegraphics[width=\textwidth]{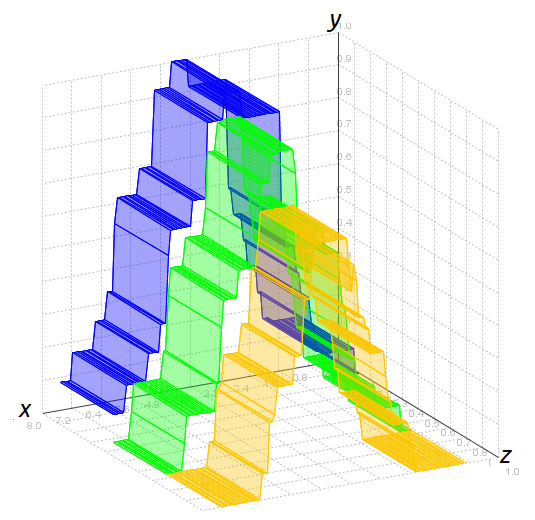}\caption{Moderately Difficult}
		\label{subfig:MD}
	\end{subfigure}
	\begin{subfigure}[b]{0.2\textwidth}
		\includegraphics[width=\textwidth]{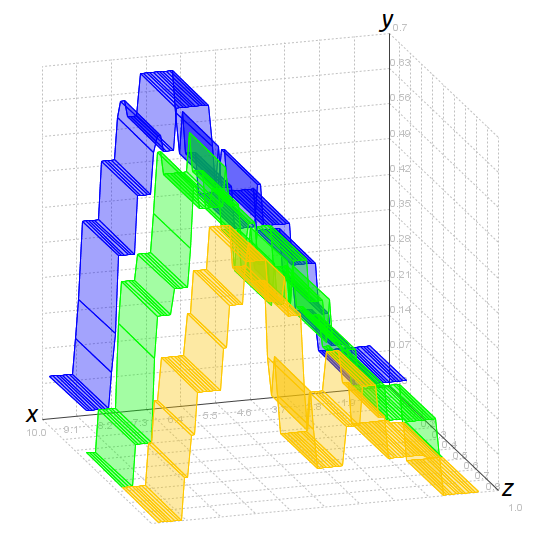}\caption{A little bit difficult}
		\label{subfig:ALBD}
	\end{subfigure}
	~
	\begin{subfigure}[b]{0.2\textwidth}
		\includegraphics[width=\textwidth]{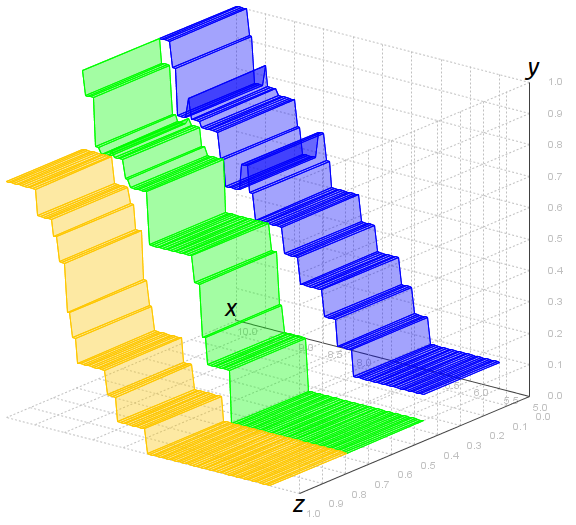}\caption{Not at all difficult}
		\label{subfig:NAAD}
	\end{subfigure}
	\caption{GT2 FSs modelling the linguistic descriptors: \textit{ITD}, \textit{ED}, \textit{MD}, \textit{ALBD}, and \textit{NAAD}, generated from Patients, Physiotherapists and Surgeons.}
	\label{fig:gt2LTs}
\end{figure}

In this last section, we consider the case of three groups of participants (i.e., patients, surgeons and physiotherapists) who provided IV responses in response to their perception about 5 linguistic descriptors used in a questionnaire on \textit{Quality of Life}. These IV responses were subsequently modelled as GT2 FSs using the IAA (see Fig. \ref{fig:gt2LTs}) where inter-group agreement and intra-respondent uncertainty are reflected through the primary and secondary memberships respectively. Then, we applied \eqref{eq:agrRatGT2} to the 5 GT2 FSs and obtained their inter-group agreement $\gamma$ levels (see Table \ref{tab:agr}). 

\begin{table}[!ht]
  \centering
  \caption{Inter-group Agreement Ratio}
    \begin{tabular}{lcccc}
    \toprule
    Linguistic Term & $\gamma$ & \textbf{Support} & \textbf{Centroid} & \textbf{Height} \\
    \midrule
    Impossible to do & 0.8324 & [0.0, 3.2] & 0.7395 & 1 \\
    Extremely difficult & 0.8322 & [0.0, 4.3] & 1.8606 & 0.9167 \\
    Moderately difficult & 0.8038 & [0.9, 7.5] & 4.3715 & 1 \\
    A little bit difficult & 0.8054 & [1.5, 9.8] & 6.4101 & 0.6667 \\
    Not at all difficult & 0.7864 & [5.5, 10] & 9.0771 & 1 \\
    \bottomrule
    \end{tabular}%
  \label{tab:agr}%
\end{table}

Looking at the individual T1 FS depictions of the groups' perceptions on the selected linguistic terms in Fig. \ref{fig:T1LTs}, it can be noticed that in all cases the different FSs are roughly distributed in the same region therefore, one could expect a high level of inter-group agreement. This can be confirmed in the results presented in Table \ref{tab:agr} where for all GT2 FS models (shown in \ref{fig:gt2LTs}) the inter-group agreement is relatively high ($>0.78$ in all cases).

\section{Conclusion} \label{conc}
In this paper we set out to develop a novel FS measure based on the previously proposed Agreement ratio \cite{Navarro2016b} for T1 FS base models of human sourced data. The aim was to extend the original measure which numerically expresses the level of agreement (denoted by the primary membership) within a group of individuals (\textit{inter}-source uncertainty) when modelled by a T1 FS. The proposed extension of the Agreement ratio considers the case of modelling the perception of groups of individuals using GT2 FSs which, as per design, can model two types of uncertainty (e.g., inter-source and inter-group) across two dimensions (related to the primary and secondary membership axis respectively), thus, providing a means for evaluating the strengths of agreement within a group of people and among groups of people.

Empirical demonstrations of the developed measures and subsequent analysis were performed by considering synthetic cases and a real world case of study. The presented analysis considered information captured through IV questionnaires, in particular from groups of people with respect to very specific matters such as their perception on LTs. In particular, the analysis considered the agreement across three groups of respondents with different background and was modelled through \textit{z}Slice Based General Type-2 Fuzzy Sets.

In general, results of the analysis on the FS agreement models of the IV TESS case study allowed to show that the agreement w.r.t 5 LTs is strong among the three groups of people. Empirical validations of the proposed measures confirmed that these follow an intuitive behaviour when comparing LTs which are notoriously more/less ambiguous than others. Thus, the results confirmed that the proposed measures can provide a means of assessing FSs generated using the IAA technique. Potential applications of the proposed measure are suggested in intercommunication scenarios where it is of interest to analyse how consistent the perception of a LT is (or a set of LTs are) among different groups of people with different background/experience. Future work needs to consider the case of using other FS modelling techniques and compare the proposed measure against other FS indexes in order to highlight its usefulness in diverse areas such as in Computing with Words \cite{Zadeh1996} applications.

\ifCLASSOPTIONcompsoc
  \section*{Acknowledgments}
\else
  \section*{Acknowledgment}
\fi

The authors would like to thank all the patients from Nottingham University Hospital (City Campus) who kindly took part in the TESS survey assisted by the Sarcoma specialists Lynsey Green (Physiotherapist) and Rob Ashford (Surgeon).



%
\bibliographystyle{myIEEEtran}
\bibliography{references.bib}

\end{document}